# Deep Causal Learning for Robotic Intelligence

Yangming Li

**Abstract**

This invited review discusses causal learning in the context of robotic intelligence. The paper introduced the psychological findings on causal learning in human cognition, then it introduced the traditional statistical solutions on causal discovery and causal inference. The paper reviewed recent deep causal learning algorithms with a focus on their architectures and the benefits of using deep nets and discussed the gap between deep causal learning and the needs of robotic intelligence.

## I. INTRODUCTION

Intelligent Robots infer knowledge about the world from sensor perception, estimate status, model the world, and plan and execute tasks. Although intelligent robots have achieved remarkable progress in the past two decades, it is still challenging to improve the reliability of intelligent robots in the real world. The challenges roots in both the wide variance of environments and robotic tasks and the uncertainties of the world, sensor observation, the models and the status, and the execution of tasks.

State of art robotic systems achieve intelligence from two types of methods. One method is to transplant knowledge, including proven theory, established models, and coded rules, to robots. This type of achieved intelligence has predictable performance, is explainable, but lacks adaptiveness, their complexity grows exponentially with respect to the complexity of tasks. The other method is to learn knowledge from observations. This type of achieved intelligence is the opposite of the previous type, as it is salable but difficult to explain.

Human beings, or even animals, learn the world effortlessly from an early age and build up prior knowledge quickly to make causal decisions in daily life. Let's use the perception of an object's physical properties as an example. Even infants demonstrate their instinctual behavior of inspecting a new toy with their hands and eyes in tandem for learning the toy's properties[1]. Robots, in comparison, still have problems understanding and operating the most commonly used objects in daily life. As clear as causality is critical from low-level visual perception to high-level decision-making, state of art robots rarely establish the causal relationship and utilize the relationship to improve intelligence. But there are examples that "causal" relationships improve robotic intelligence. For example, Simultaneous Localization and Mapping (SLAM) explicitly utilizes the fact that causal relationships of robot movements causally changing observations and use a Bayesian network to improve mapping and localization simultaneously[2].

Causal learning consists of the causal discovery and causal inference. Classical causal learning methods. Causal discovery learns cause-effect relationships, and causal inference estimates how much the changes in factors impact other factors. Traditional causal learning algorithms mainly use statistical theories and tools. With the development of deep learning technology, there are trends that use deep learning to improve causal learning with high-dimensional data and big data and trends that use causal learning to improve deep learning model expandability, extrapolation capability, and explainability. Although these emerging techniques are not developed and tested on intelligent robots yet, they do have great potential to improve robotic intelligence and expand the applicability of intelligent robots. This paper incompletely but systematically reviews causal cognition, causal learning, and deep causal learning, and discusses the need for deep causal learning in robotic intelligence. The rest of the paper is organized as follows: Section 2 briefly introduces causal cognition from the psychological perspective, section 3 presents the statistical causal discovery and causal inference, section 4 discusses deep causal learning for robotic intelligence, and the last section concludes the paper.

## II. CAUSAL COGNITION AND INTELLIGENCE

Regardless of the debut on cognition mechanism, modern physiological studies generally support that human subjects cognize causal regularities is more sophisticated than that of any other animal on the planet[3]. Causal cognition has major differences with associative learning, as it can improve inferences from nonobvious and hidden causal relationships[4]. Learning causal relations is critical to human beings [4, 5], as it confers an important advantage for survival[6, 7].

It is clear that to achieve or avoid an outcome, one may want to predict with what probability an effect will occur given that a certain cause of the effect occurs. It remains mysterious as to how the input noncausal empirical observations of cues and outcomes yield output values. Studies have shown that causal cognition emerges early in development[6]. Researchers are amazed by how children learn so much about the world so quickly and effortlessly[8, 9]. Studies have demonstrated that infants, as young as 4.5 months, register particular aspects of physical causality[5, 10–12], toddlers recognize various causal relations in the psychological domain, especially about others' desires and intentions[13, 14], and preschoolers understand that biological and psychological events can rely on nonobvious, hidden causal relations [15, 16]. Adults use substantive prior knowledge about everyday physics and psychology to make new causal judgments[17].

III. Causal Learning

As presented in the previous section, causal learning is associated with human intelligence and is widely studied. Traditional causal learning uses statistical methods to discover knowledge from data and perform causal inference. These methods are widely used in the field of medical science, economics, epidemiology, etc, but rarely used in the domain of intelligent robotics[18, 19].

A. Causal Discovery

Causal discovery learns the causal structure that represents the causality between observations X, treatments t, and outcomes y.

Traditional causal discovery relies on statistically verifying potential causal relationships or estimating functional equations to establish causal structures. Generally, there are four types of representative algorithms for traditional causal discovery: the constraint-based algorithms and the score-based algorithms, which rely on statistical verification, the functional causal model-based algorithms, which rely on functional estimation, and the hybrid algorithms, which fuse multiple algorithms[18].

*1) constraint-based algorithms:* Constraint-based algorithms analyze conditional independence in observation data to identify causal relationships. This family of algorithms often uses statistical testing algorithms to determine the conditional independence of two variables, given their neighbor nodes, then further determine the direction of the causality.

Mathematically, we can use three variables $X$, $Y$, and $Z$ to explain Constraint-based algorithms. The causal relationship is verified by conditional independence, for example, $X \perp Y \mid Z$, which is equivalent to zero conditional information $I[X; Y \mid Z] = 0$. This is defined as Faithfulness in causal learning, as explained in Definition 3.1. If the three variables are discrete, $\chi^2$ and $G^2$ can verify the conditional independence based on the contingency table of $X$, $Y$, and $Z$. If the three variables are linear and multivariate Gaussian, we can verify the conditional indecency by test if the partial correlation is zero. For other circumstances, it often needs extra assumptions to ensure the verification is computationally tractable.

*Definition 3.1:* (Faithfulness). Conditional independence between a pair of variables, $x_i \perp x_j \mid x^-$ for $x_i \neq x_j$, $x^- \subseteq X \backslash \{x_i, x_j\}$, can be estimated from a dataset X iff $x^-$ d-separates $x_i$ and $x_j$ in the causal graph $G = (V, E)$.

The conditional independence is symmetric, and additional tests are required to determine the orientations of edges. When $X \perp Y \mid Z$, there are three possible graphical structures, including two chains ($X \leftarrow Z \leftarrow Y$ and $X \rightarrow Z \rightarrow Y$) and a fork $X \leftarrow Z \rightarrow Y$. The determination of which structures are induced based on the adjacency among variables, the background knowledge, etc. When $X \not\perp\!\!\!\perp Y \mid Z$ it is a collision structure ($X \rightarrow Z \leftarrow Y$).

Constraint-based algorithms used assumptions to improve efficiency and effectiveness for causal discovery from data. For example, the Peter-Clark algorithm assumes i.i.d. sampling and no latent confounders, which prunes edges between variables by testing conditional independence based on observations data, and determines and propagates the orients to form the directed acyclic graph (DAG)[20]. The inductive causation algorithm assumes stable distributions (Definition 3.2), tests conditional independence to find the associative relationship between variables, finds collision structures, determines orients based on variable's adjacency, and propagates directions[21].

*Definition 3.2:* (Stable Distribution). a distribution is stable if a linear combination of two independent random variables with this distribution has the same distribution, up to location and scale parameters.

Other constraint-based algorithms aim to relax the assumptions and extend the causal discovery to other families of distributions [22–24], causal discovery from data with unobserved confounders [18, 25].

*2) score based algorithms:* Score-based algorithms learn causal graphs by maximizing the goodness-of-fit test scores of the causal graph $G$ given observation data $X$[20]. Because these algorithms replaced the conditional independence tests with the goodness-of-fit tests, they relax the assumption of faithfulness (Definition 3.1) but often increase computational complexity. This is because the scoring criterion $S(X, G)$ enumerates and scores the possible graphs under parameter adjustments. For example, the popular Bayesian Information Criterion adopts the score function $S(X, G) = \log P(X|\hat{\theta}, G) - J/2 \log(n)$ to find the graph that maximizes the likelihood of observing the data, while the number of parameters and the

sample size is regularized, where $\hat{\theta}$ is the Maximum Likelihood Estimation of the parameters and $J$ and $n$ denote the number of variables and the number of instances, respectively[26].

It is not tractable to score all possible causal graphs given observation data because it is NP-hard[27] and NP-complete [28]. In practice, score-based algorithms use heuristics to find a local optimum[29, 30]. For example, the Greedy Equivalence Search algorithm uses Bayesian Dirichlet equivalence score $S_{BD}$:

$$S_{DB}(X, G) = \log \prod_{j=1}^{J} 0.001^{(r_j - 1)q_j} \prod_{k=1}^{q_j} \frac{\Gamma(10/q_j)}{\Gamma(10/q_j + N_{jk})} \prod_{l=1}^{r_j} \frac{\Gamma(10/r_j/q_j + N_{jkl})}{\Gamma(10/r_j/q_j)}, \tag{1}$$

to score a graph $G$, where $r_j$ and $q_j$ signify the numbers of configurations of variable $x_j$ and the numbers of configurations of parent set $Pa(x_j)$, respectively, $\Gamma(\cdot)$ denotes the Gamma function, and $N_{jkl}$ denotes the number of records of $x_j = k$ and $Pa(x_j)$ are in the $k$-th configuration.

Widely used score-based algorithms optimize the searching and the scoring process based on the assumptions such as linear-Gaussian models[31], discrete data [32], and sparsity [33]. There are also works on relaxing the assumptions for causal discovery from nonlinear and arbitrarily distributed data [34]. Compared with constraint-based algorithms, score-based algorithms can compare the output models in the space searched for model selection.

*3) Functional Causal Models based algorithms:* Functional Causal Models based algorithms represent the causal relationship with functional equations (Define 3.3).

*Definition 3.3:* (Functional Equation). a functional equation represents a direct causal relation as $y = f_\theta(X, n)$, where $X$ is the variables that directly impact the outcome $y$, $n$ is noise with $n \perp X$, and $f_\theta$ is the general form of a function.

Causal discovery with functional equations can be expressed as sorting causal orders (which variables depend on which) from observation data. We use Linear Non-Gaussian Acyclic Model to explain the process with a simple linear case $x = Ax + \mu$, where x denotes the variable vector, A denotes the adjacency matrix, and $\mu$ denotes the noise independent of x. With this representation, the causal discovery is the equivalent of estimating a strictly lower triangle matrix $A$ that determines the unique causal order $k(x_i)$, $\forall x_i \in mathbfx$, which can be performed in the form of Matrix permutation as described in [35].

Functional Causal Models based algorithms have demonstrated effectiveness in producing unique causal graphs. For example, the post-nonlinear causal model learns the causal relationship that can be represented by a post-nonlinear transformation on a nonlinear effect of the cause variables and additive noises[36]. This algorithm can be further improved with independent component analysis [37] and relaxed by warped Gaussian process with the noise modeled by the mixture of Gaussian distributions[38]. Compared to the constraint-based algorithms and the score-based algorithms, Functional Causal Models based algorithms are able to distinguish between different DAGs from the same equivalent class.

*4) hybrid algorithms:* Hybrid algorithms combine multiple algorithms to overcome problems that exist in constraint-based algorithms or score-based algorithms. For example, [39] uses the Max-Min Parents and Children algorithm (constrained-based) to learn the skeleton of the causal graph and uses Bayesian scoring hill-climbing search (score-based) to determine the orients of edges. [40] uses the conditional independence test to learn the skeleton of the causal graph and use a metric to search good network structures.

### B. Causal Inference

Causal inference is the process of estimating the changes of outcomes y given treatments t. Before we discuss causal inference algorithms, let's define the metrics (Definition 3.4) for measuring causal inference. ATE, ATT, CATE, and ITE measure the treatment effects at the population, treated group, a subgroup of a given feature $x$, and individual levels, respectively.

*Definition 3.4:* (Treatment Effect).

- Average Treatment Effect (ATE): $\text{ATE} = E[Y(w = 1) - Y(w = 0)]$.

- Average Treatment Effect on Treated Group (ATT): $\text{ATT} = E[Y(w = 1) \mid w = 1] - E[Y(w = 0) \mid w = 1]$.

- Conditional Average Treatment Effect (CATE): $\text{CATE} = E[Y(w = 1) \mid X = x] - E[Y(w = 0)|X = x]$.

- Individual Treatment Effect (ITE): $\text{ITE}_i = Y_i(w = 1) - Y_i(w = 0)$.

Causal inference estimates the treatment effects for specific groups. However, the different distributions of groups and the existence of confounders make the task very challenging. According to the methodological differences, existing classical algorithms for addressing these problems can be grouped into Re-weighting based algorithms, stratification-based algorithms, batching-based algorithms, and tree-based algorithms.

*1) Re-weighting based algorithms:* Re-weighting-based algorithms assign appropriate weights to the samples to create pseudo populations or re-weight the covariates to mitigate the differences in the distributions between the treated groups and the control groups. These algorithms are designed to address the selection bias between the treated groups and the control groups.

Both samples and covariate re-weighting are used to address the selection bias. The inverse propensity weighting algorithm is one of the pioneering works on re-weighting samples. This algorithm uses the Propensity Scores (Definition 3.5) to find the appropriate weights for samples as $r = T/e(x) + (1 - T)/(1 - e(x))$, where $T$ is the treatment.

*Definition 3.5:* (Propensity Score). Propensity Score $e(x)$ is the conditional probability of assignment to a particular treatment given a vector of observed covariates $e(x) = Pr(T = 1|X = x)$.

With the re-weighting, the ATE is defined as: $A\hat{T}E = 1/N_T \sum_{i=1}^{N_T} T_i Y^F/\hat{e}(x) - 1/N_C \sum_{i=1}^{N_C} (1 - T_i) Y^F/(1 - \hat{e}(x))$. This method is sufficient to remove bias, however, heavily relies on the correctness of propensity scores[41]. Along the lines of propensity score-based sample re-weighting, the Doubly Robust Estimator combines the propensity score weighting with the outcome regression to remain unbiased as long as the propensity score or outcome regression is correct(Fig. 1)[41], the overlap weights algorithm down-weighting the units in the tails of the propensity score distribution to emphasizes the target population with the most overlap in observed characteristics to overcome the extreme propensity score problem [42].

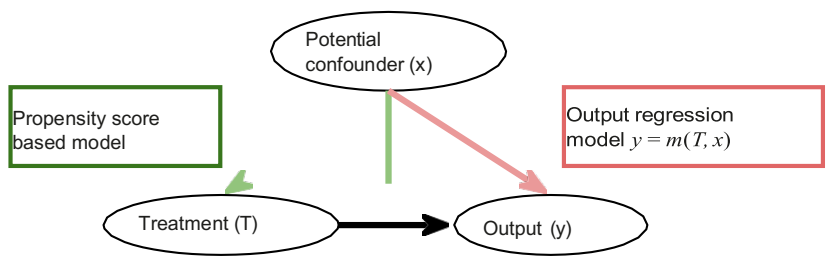

Fig. 1: Doubly Robust Estimator.

The covariate re-weight algorithms learn sample weights from data through regression. To re-weight covariate, [43] uses a maximum entropy re-weighting scheme to calibrate sample weights to match the moments of the treated group and the control group and minimizes information loss by keeping weights close to the base weights.

There are also algorithms that balance distributions with both covariate and sample re-weighting. Covariate balancing propensity score estimates the propensity score by solving: $E[W_i \tilde{x}_i/e(x_i; \beta) + (1 - W_i)\tilde{x}_i/(1 - e(x_i; \beta))]$ to measure the probability of being treated and covariate balancing score and improves the empirical performance of propensity score matching[44]. Data-Driven Variable Decomposition (D2VD) balances distribution by automatically decomposing observed variables confounders, adjusted variables, and irrelevant variables[45], Differentiated Confounder Balancing (DCB) selects and differentiates confounders, and re-weighting both the sample and the confounders to balance distributions [46].

*2) Stratification based algorithms:* Stratification-based algorithms split observation into subgroups, which are similar under certain measurements. With subgroups that have balanced distributions, ATE is estimated as $\tau^{\text{strat}} = \sum_{j=1}^{J} q(j)\overline{Y}_t(j) - \overline{Y}_c(j)$. For example, if a model can predict the strata in which subjects always stay in the study regardless of which treatment they were assigned, then the data from this strata is free of selection biases[47, 48]. The stratification can be performed on samples on the basis of the propensity score to improve the estimation robustness, as explained in the marginal mean weighting through stratification algorithm [49]. The stratification algorithms can also be combined with propensity score-based algorithms as a prepossessing of data to remove imbalances of pre-intervention characteristics [50].

*3) Matching based algorithms:* Matching-based algorithms use specific distance measurements to match samples in the treatment group with ones in the control group to estimate the counterfactuals and reduce the estimation bias of confounders. Matching-based algorithms require the definition of distance metrics and the selection of matching algorithms. Euclidean distances and Mahalanobis distances are commonly used as distance metrics in the original data space, while transformations, such as propensity score-based transformation, and observed outcome information are commonly used in the transformed feature space [19, 51]. For matching algorithms, Nearest Neighbor, Caliper, Stratification, and Kernel-based methods are all widely adopted[18]. It is worth noticing that matching-based algorithms can be used in data selection, as well as experimental design and performance. The latter uses matching to identify subjects whose outcomes should be collected [52, 53], which potentially reduces costs and difficulty in collecting effective data.

*4) Tree-based methods:* A tree structure naturally divides data into disjoint subgroups. While the subgroups have similar $e(x)$, the estimation of the treatment effect is unbiased. Bayesian Additive Regression Trees (BART), a Bayesian "sum-of-trees" model, is a flexible approach to fitting a variety of regression models while avoiding strong parametric assumptions.

With BART, the treatment $y$ is the sum of subgroups as $y = g(x; T_1, \theta_1) + \cdots + g(x; T_n, \theta_n) + \sigma$, where $\sigma$ is Gaussian

White noise[54]. Similarly, the Classification And Regression Trees (CART) algorithm also splits data into classes that belong to the response variable. Being different from BART, CART recursively partitions the data space and fits a simple prediction model for each partition[55]. Causal Forests ensemble multiple causal trees to achieve a smooth estimation of CATE. Causal Forests are based on Breiman's random forest algorithm and maximize the difference across splits in the relationship between an outcome variable and a treatment variable for revealing how treatment effects vary across samples[56].

## IV. Deep Causal Learning for Robotic Perception

### A. Deep Causal Learning

Deep learning (DL) successfully attracted researchers from all fields as DL demonstrated the power and the simplicity of learning from data[57]. The majority of existing DL algorithms use specialized architecture to establish end-to-end relationships from observation data, for example, Convolutional Neural Networks (CNN) for data with spatial locality, Recurrent Neural Networks (RNN) for data with sequential or temporal structure, Transformers for data with context information, Autoencoders for data need compressed representation, Generative Adversarial Networks for data need domain adaption[58–62]. Despite the remarkable success DL achieved, some challenges remain in DL, such as model expandability, extrapolation capability, and explainability. Causal Learning (CL), on the other hand, discovers knowledge, explains prediction, and has extendable structures, but struggles with high dimensional data and scalability problems. It is encouraging to compliment DL with CL and vice versa. Actually, recent studies make great progress and demonstrated Causal Deep Learning has advantages as they can use prior knowledge to disentangle modeling problems and reduce data needs[63–65], have superior performance on extrapolating unseen data [66, 67], modularize learning problems, incrementally learn from multiple studies[68–70], and demonstrate its potential as a solution to artificial general intelligence [71–73].

Below, we introduce some of the representative works in plain language, with a focus on the network architecture and the benefits of using the architecture. Because the algorithms we reviewed share many common characteristics, such as most of them use two or more neural networks, most of the representation learning involves CNN, etc., we categorize the algorithms into the following categories to maximize the uniqueness among the categories.

*1) Using DL for learning representation:* Balancing Neural Networks and Balancing Linear Regression is one of the pioneer works that use deep neural networks to solve the problem of causal learning from high dimensional data[74]. These algorithms learn a representation $g : X \to \mathbb{R}^d$ through deep neural networks or feature re-weighting and selection, then based on the features $g(X)$ learn the causal effect $h : \mathbb{R}^d \times T \to \mathbb{R}$. These models learn balanced representations that have similar distributions among the treated and untreated groups and demonstrated effectiveness in cases that have one treatment.

Similarity preserved Individual Treatment Effect (SITE) uses two networks to preserve local similarity and balances data distributions simultaneously[75]. The first network is a representation network, which maps the original pre-treatment covariate space $X$ into a latent space $Z$. The second network is a prediction network, which predicts the outcomes based on the latent variable $Z$. The algorithm uses Position-Dependent Deep Metric and Middle Point Distance Minimization to enforce two special properties on the latent space Z, including the balanced distribution and preserved similarity. Adaptively similarity-preserved representation learning method for Causal Effect estimation (ACE) preserves similarity in representation learning in an adaptive way for extracting fine-grained similarity information from the original feature space and minimizes the distance between different treatment groups as well as the similarity loss during the representation learning procedure[76]. ACE applied Balancing and Adaptive Similarity preserving (BAS) regularization to the representation space. The BAS regularization consists of distribution distance minimization and adaptive pairwise similarity preserving, therefore, decreasing the ITE estimation error.

[77] presented a theory and an algorithmic framework for learning to predict outcomes of interventions under shifts in design—changes in both intervention policy and feature domain. This framework combines representation learning and sample re-weighting to balance source and target designs, emphasizing information from the source sample relevant to the target. As the result, this framework relaxes the strong assumption of having a well-specified model or knowing the policy that gave rise to the observed data.

*2) End-to-end deep causal inference:* [78] Treatment-Agnostic Representation Networks (TARNET) proposed to estimate ITE based on the Rubin potential outcomes framework under the assumption of strong ignorability. This algorithm uses Integral Probability Metrics to measure distances between distributions and derives explicit bounds for the Wasserstein and Maximum Mean Discrepancy (MMD) distances. Therefore, this algorithm is an end-to-end regularized minimization procedure that fits the balanced representation of the data and a hypothesis for the outcome. Based on this work, [79] proposed a context-aware importance sampling re-weighing scheme to estimate ITEs, which addresses the distributional shift between the source (outcome of the administered treatment, appearing in the observed training data) and target

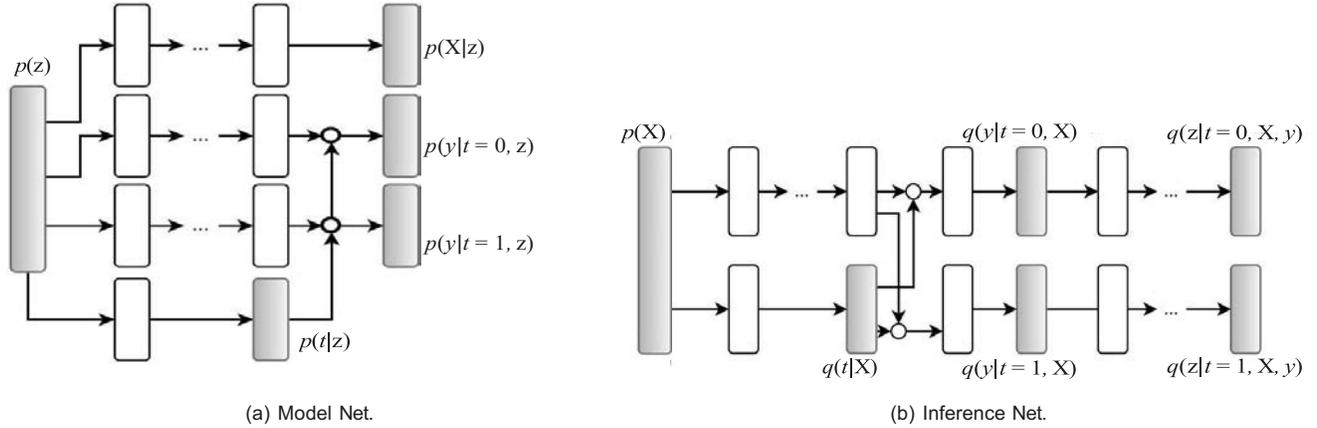

(a) Model Net.

(b) Inference Net.

Fig. 2: Causal Effect Variational Autoencoder[82].

(outcome of the alternative treatment) that exists due to selection bias. Perfect Matching augments samples within a minibatch with their propensity-matched nearest neighbors to improve inference performance in settings with many treatments[80]. Perfect Matching is compatible with other architectures, such as the TARNET architecture, and extends these architectures to any number of available treatments. Perfect Matching also uses the nearest neighbor approximation of Precision in the Estimation of Heterogenous Effect with multiple treatments to select models without requiring access to counterfactual outcomes.

[81] models the inference of individualized causal effects of a treatment as a multitask learning problem. The algorithm uses a propensity network and a potential outcomes network to estimate ITE (Definition 3.4). The propensity network is a standard feed-forward network and is trained separately to estimate the propensity score $e(x_i)$ (Definition 3.5) from $(x_i, t_i)$. Through assigning "simple models" to subjects with very high or very low propensity scores ($e(x_i)$ close to 0 or 1), and "complex models" to subjects with balanced propensity scores ($e(x_i)$ close to 0.5), it alleviates the selection bias problem. The potential outcomes network is a multitask network that models the potential outcomes $E[Y^{(1)}|x^i]$ and $E[Y_i^{(0)}|x^i]$ as two separate but related learning tasks, therefore, the treatment assignments and the subjects' characteristics are fully utilized.

*3) Autoencoder based algorithms:* Causal Effect Variational Autoencoder (CEVAE) uses Variational Autoencoders (VAE) structures to estimate individual treatment effects[82]. The algorithm uses an inference network (Fig. 2b) and a model network (Fig. 2a) to simultaneously estimate the unknown latent space summarizing the confounders and the causal effect, based on latent variable modeling. Because the algorithm uses the two networks to utilize both the causal inference with proxy variables and latent variable modeling, its performance is competitive with the state-of-the-art on benchmark datasets and has improved robustness on the problems with hidden confounders.

The Deep-Treat algorithm uses two networks for constructive effective treatment policies by addressing the problems of the observed data being biased and counterfactual information being unavailable[83]. The first network is a bias-removing auto-encoder, which allows the explicit trade-off between bias reduction and information loss. The second network is a feedforward network, which constructs effective treatment policies on the transformed data.

Task Embedding based Causal Effect Variational Autoencoder (TECE-VAE) scales CEVAE with task embedding for estimating individual treatment effect using observational data for the applications that have multiple treatments[84]. TECE-VAE also adopts the Encoder-Decoder architecture. The encoder network takes input $X$ to generate distribution for $z$. The decoder network uses $z$ to reconstruct features $X$, treatments $t$, and outcomes $y$. TECE-VAE uses information across treatments and is robust to unobserved treatments.

The Conditional Treatment-Adversarial learning based Matching method (CTAM) uses treatment-adversarial learning to effectively filter out the nearly instrumental variables for processing textual covariates[85]. CTAM learns the representations of all covariates, which contain text variables, with the treatment-adversarial learning, then performs the nearest neighbor matching among the learned representations to estimate the treatment effects. The conditional treatment adversarial training procedure in CTAM filters out the information related to nearly instrumental variables in the representation space, therefore, the treatment discriminator, the representation learner, and the outcome predictor work together in the adversarial learning way to predict the treatment effect estimation with text covariates. To be more specific, the treatment discriminator is trained to predict the treatment label, while the representation learner works with the outcome predictor for fooling the treatment discriminator.

Reducing Selection Bias-net (RSB-net) proposed to use two networks to address the selection bias problem[86]. The first net is an auto-encoder that learns the representation. This auto-encoder uses a Pearson Correlation Coefficient (PCC) based on regularized loss and explicitly differentiates the bias variables with the confounders that affect treatments and outcomes and the variables that affect outcomes alone. Therefore, the confounders and the variables affecting outcomes are fed into the second network, which uses the branching structure network to predict outcomes.

Variational Sample Reweighting (VSR) algorithm uses a variational autoencoder to remove the confounding bias in the applications with bundle treatments [87]. VSR simultaneously learns the encoder and the decoder by maximizing the evidence lower bound.

*4) Generative Adversarial Nets-based algorithms:* Generative Adversarial Nets for inference of Individualized Treatment Effects (GANITE), as suggested by the name, inferring ITE based on the Generative Adversarial Nets (GANs) framework[88]. The algorithm uses a counterfactual generate, G, to generate potential outcome vector $y^~$ based on features X, treatments $t$, and factual outcome $y_f$. Then the generated proxies are passed to an ITE generator that generates potential outcome $y^\hat$ based on feature X. As the Generative Adversarial Nets[62], GANITE uses a discriminator for G, $D_G$, and a discriminator for I, $D_I$ to boost the training performance for the generators. $D_G$ maps pairs $(X, y^-)$ to vectors in $[0, 1]^k$ with the $i$−th component to representing the probability that the $i$−th component of $y^~$ is the factual outcome. Similarly, $D_I$ maps a pair $x, y^*$ to $[0, 1]$ representing the probability of $y^*$ being from the data $\tilde{D}$.

Causal Effect Generative Adversarial Network (CEGAN) utilizes an adversarially learned bidirectional model along with a denoising auto-encoder to address the confounding bias caused by the existence of unmeasurable latent confounders[89]. CEGAN has two networks, a prediction network (consisting of a generator, a prediction decoder, an inference net, and a discriminator), and a reconstruction network (a denoising autoencoder, whose encoder is used as the generator in the prediction network).

SyncTwin proposed to construct a synthetic twin that closely matches the target in representation to exploit the longitudinal observation of covariates and outcomes[90]. SyncTwin uses the sequence-to-sequence architecture with an attentive encoder and an LSTM decoder to learn the representation of temporal covariates, then it constructs a synthetic twin to match the target in representations for controlling estimation bias. The reliability of the estimated treatment effect can be assessed by comparing the observed and synthetic pre-treatment outcomes.

Generative Adversarial De-confounding (GAD) algorithm estimates outcomes of continuous treatments by eliminating the associations between covariates and treatments[91]. GAD first randomly shuffles the value of covariate X into X' in order to ensure $X' \perp T$, where $T$ is the treatments. Second, GAD re-weighting samples in X so the distribution of X is identical to X'. GAD then eliminated the confounding bias induced by the dependency between $T$ and X.

Adversarial Balancing-based representation learning for Causal Effect Inference (ABCEI) uses adversarial learning to balance the distributions of covariates in the latent representation space to estimate the Conditional Average Treatment Effect (CATE)[92]. ABCET uses an encoder that is constrained by a mutual information estimator to minimize the information loss between representations and input covariates to preserve highly predictive information for causal effect inference. The generated representations are used for discriminator training, mutual information estimation, and prediction estimation.

*5) Recurrent Neural Networks-based algorithms:* Recurrent Marginal Structural Network (RMSM) uses recurrent neural networks to forecast a subject's response to a series of planned treatments[93]. RMSM uses the encoder-decoder architecture. The encoder learns representations for the subject's current state by using a standard LSTM to predict one-step-ahead outcome $(Y_{r+2})$ given observations of covariates and actual treatments. The decoder forecast treatment responses on the basis of planned future actions by using another LSTM to propagate the encoder representation forwards in time.

Counterfactual Recurrent Network (CRN) uses recurrent neural network-based encoder-decoder to estimate treatment effects over time [94]. The encoder uses domain adversarial training to build balancing representations of the patient history for maximizing the loss of the treatment classifier and minimizing the loss of the outcome predictor. The decoder updates the outcome predictor to predict counterfactual outcomes of a sequence of future treatments.

Time Series Deconfounder recurrent neural network architecture with multitask-output for leveraging the assignment of multiple treatments over time and enabling the estimation of treatment effects in the presence of multi-cause hidden confounders[95]. The algorithm takes advantage of the patterns in the multiple treatment assignments over time to infer latent variables that can be used as substitutes for the hidden confounders. It first builds a factor model over time and infers latent variables that render the assigned treatments conditionally independent; then, it performs causal inference using these latent variables that act as substitutes for the multi-cause unobserved confounders.

*6) Transformer-based algorithms:* CETransformer uses transformer-based representation learning to address the problems of selection bias and unavailable counterfactual[96]. CETransformer contains three modules, including a self-supervised Transformer for representation learning which learns the balanced representation, a Discriminator network for adversarial learning to progressively shrink the difference between treated and control groups in the representation space, and outcome prediction that uses the learned representations to estimate all potential outcome representations.

*7) Multiple-branch networks and subspaces:* Dose Response Network (DRNet) uses neural networks to estimate individual dose-response curves for any number of treatments with continuous dosage parameters[97]. DRNet consists of shared base layers, $k$ intermediary treatment layers, and $k * E$ heads for the multiple treatment setting, where $k$ denotes the number of treatments and $E$ defines the dosage resolution. The shared base layers are trained on all samples, the treatment layers are only trained on samples from their respective treatment category, and a head layer is only trained on samples that fall within its respective dosage stratum.

Disentangled Representations for CounterFactual Regression (DR-CFR) proposed to disentangle the learning problem by explicitly identifying three categories of features, including the ones that only determine treatments, the ones that only determine outcomes, and the confounders that impact both treatments and outcomes[98]. Three representation learning networks are trained to identify each of the three categories of factors, and the identified factors are fed into two regression networks for identifying two types of treatments and two logistic networks to model the corresponding behavior policy.

Decomposed Representations for CounterFactual Regression (DeR-CFR) proposed to disentangle the learning problem by explicitly dividing covariants into instrumental factors, confounding factors, and adjustment factors [99]. DeR-CFR has three decomposed representation networks for learning the three categories of latent factors, respectively, has three decomposition and balancing regularizers for confounder identification and balancing of the three categories of latent factors, and has two regression networks for potential outcome prediction.

Neural Counterfactual Relation Estimation (NCoRE) explicitly models cross-treatment interactions for learning counterfactual representations in the combination treatment setting[100]. NCoRE uses a novel branched conditional neural representation and consists of a variable number of shared base layers with k intermediary treatment layers which are then merged to obtain a predicted outcome. The shared base layers are trained on all samples and serve to model cross-treatment interactions, and the treatment layers are only trained on samples from their respective treatment category and serve to model per-treatment interactions.

Single-cause Perturbation (SCP) uses a two-step procedure to estimate the multi-cause treatment effect[101]. The first step augments the observational dataset with the estimated potential outcomes under single-cause interventions. the second step performs covariate adjustment on the augmented dataset to obtain the estimator.

[102] presented three end-to-end learning strategies to exploit structural similarities of an individual's potential outcomes under different treatments to obtain better estimates of CATE in finite samples. The three strategies regularize the difference between potential outcome functions, reparametrize the estimators, and automatically learn which information to share between potential outcome functions.

Deep Orthogonal Networks for Unconfounded Treatments(DONUT) proposes a regularizer that accommodates unconfoundedness as an orthogonality constraint for estimating ATE[103]. The orthogonality constraint is defined as $< Y(t) - f(X, t) >, T - E[T \mid X = x])$, where $< \cdot, \cdot >$ is the inner product.

Subspace learning-based Counterfactual Inference (SCI) proposed to learn in a common subspace, a control subspace, and a treated subspace to improve the performance of estimating causal effect at the individual level[104]. SCI learns the control subspace to investigate the treatment-specific information for improving the control outcome inference, learns the treated subspace to retain the treated-specific information for improving the estimation of treated outcomes, and learns common subspace to share information between the control and treated subspaces for extracting the cross-treatment information and reducing the selection bias.

*8) Combining DL with statistical regulators and kernels:* Varying Coefficient neural Network (VCNet) proposed a neural network and a targeted regularization to estimate the average dose-response curve for continuous treatment and to improve finite sample performance[105].

[106] proposed the Dragonnet to exploit the sufficiency of the propensity score for estimation adjustment, and proposed the targeted regularization to induce a bias towards models. The Dragonnet uses a three-headed architecture to provide an end-to-end procedure for predicting propensity score and conditional outcome from covariates and treatment information. The targeted regularization introduces a new parameter and a new regularization term to achieve stable finite-sample behavior and strong asymptotic guarantees on estimation.

Deep Kernel Learning for Individualized Treatment Effects (DKLITE2) proposed a deep kernel regression algorithm and posterior regularization framework to avoid learning domain-invariant representations of inputs[107]. DKLITE2 works in a feature space constructed by a kernel function to exploit the correlation between inputs and uses a neural network to encode the information content of input variables.

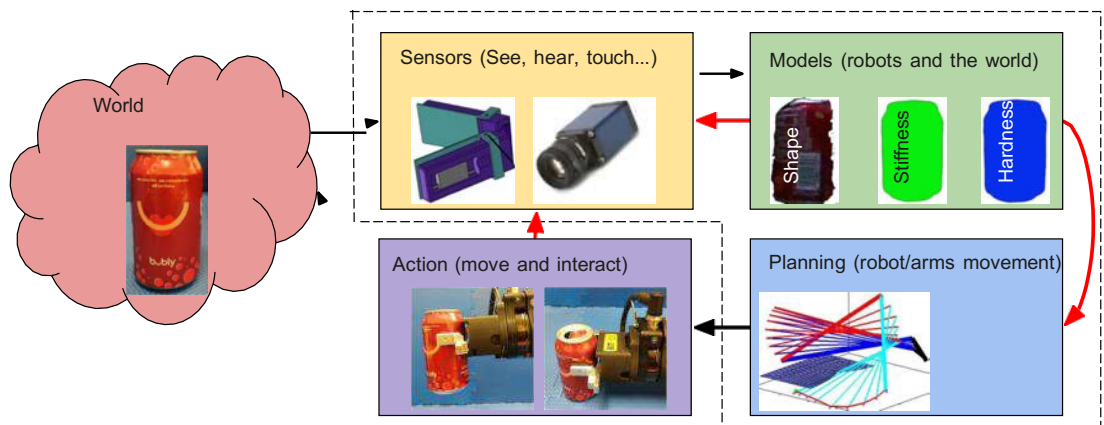

Fig. 3: Intelligent robots observe to learn, plan for interaction, and revise to improve.

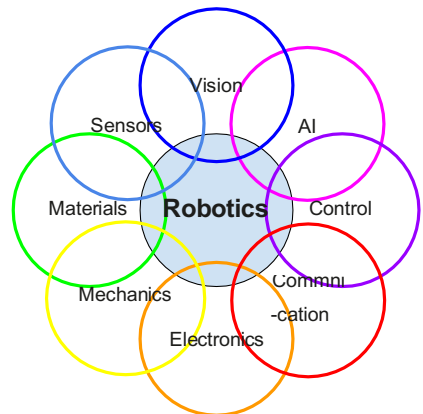

Fig. 4: Robotics is multidisciplinary.

From the above introduction, it is clear that DL architectures are widely used in CL for reducing dimensionality, processing temporal data, balancing distributions, and removing confounding bias and selection bias. Among the architectures, autoencoder and GAN are particularly popular. From the application perspective, most of the above works focus on estimating ITE, and the works on estimating ATE and CATE do exist.

### B. Robotic Intelligence

Intelligent robots observe and interact with the environment. They use various sensors to achieve information[1], use models to represent the world and estimate their status[108], plan motions[109], execute the planning and correct the execution to achieve the tasks[110], under environmental uncertainties, sensory noises, modeling uncertainties, and execution errors(Fig. 3)[111].

Robotics is multidisciplinary and widely involves technology from various fields (Fig. 4). This fact causes the fact that existing intelligent robot research mainly focuses on specific robotic technology[112, 113]. To improve the application of intelligent robots, we need not only the improvement on a specific technology, such as AI, also the integration of robotic techniques and the adoption of domain knowledge, which is essential to the real-world applications[114–116]. However, domain knowledge is even out of the robotics field and state of the art deep learning alone can not bridge the gap because of the scarcity of data, etc [117–119]. We will use three examples, a low-level visual tracking example, a middle-level motion planning, and a high-level task planning to illustrate why we believe that deep causal learning has the potential to bridge the gap.

Visual tracking is an important problem in robotics and is widely studied in computer vision, AI, and the robotics field (Table I). Particularly for visual tracking in endoscopic surgeries, there are a large number of results that can address challenges of illumination changes, occlusions, lens blur, drastic scene changes, deformation, etc.[120–123]. However, visual tracking remains challenging in endoscopic surgeries because all these adverse factors exist simultaneously and deteriorate tracking performance. Meanwhile, these adverse factors, along with the variance of pathology and anatomy, make the need for training data grow beyond our capacity. Therefore, we believe that deep causal learning is needed to disentangle the problems for robots.

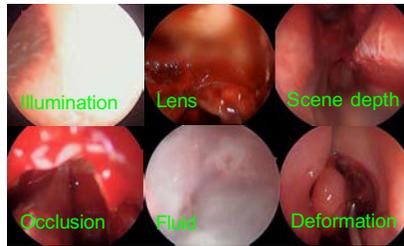

Fig. 5: Visual tracking in endoscopic surgery.

| | Static | Dynamic object | Low texture | Image quality | Illumi-nation | Reco-very | Mo-tion | Defor-mation | Scene depth |
|---|---|---|---|---|---|---|---|---|---|
| sparse | [124, 125] | [126–129] | [130, 131] | [132] | [131, 133] | [134, 135] | [136] | | |
| semidense | [137, 138] | [139] | [140] | [137] | | | | | |
| full-dense | [141] | [142–144] | [145–148] | [149, 150] | [151–153] | | [147] | [154, 155] | [147, 156] |

TABLE I: Visual tracking algorithms for addressing various challenges.

Motion planning is widely studied in robotics [157, 158]. However, in real-world applications, intelligent robots not only need to know where they move to and how to move there but also need to know if there are other application-specific requirements. For example, it is well-studied that movement patterns impact surgical outcomes (Fig. 6)[114, 116], but it is not trivial to plan motions for a robot for various treatment procedures[110, 111, 159, 160]. Therefore, we believe that deep causal learning, which naturally uses graphical structures to represent knowledge, can effectively incorporate domain knowledge with robotic techniques.

Task-level planning involves multiple decision-making and is specific to applications. For example, robotic surgery, as one of the most successful real-world applications of robotic technology, is still fully teleoperated, despite literature found that many surgical accidents were caused by wrong operations of surgical robots[161–163]. While we do believe there are legal and regulatory barriers that prevent the adoption of autonomous technology, we argue that the main problem is that we lack the technology to handle environmental and task variance. For example, robots have problems dynamically adapting to changes and determining the completeness of surgery[164].

## V. Conclusion

Recently Deep Causal Learning has demonstrated its capability of using prior knowledge to disentangle modeling problems and reduce data needs, improve performance on extrapolating unseen data, modularize learning problems, and incrementally learn from multiple studies. Inspired by these new findings, this work incompletely, but systematically discusses causal cognition, statistical causal learning, deep causal learning, and the need for deep causal learning in intelligent robots, and argues that deep causal learning is the new frontier for intelligent robot research.

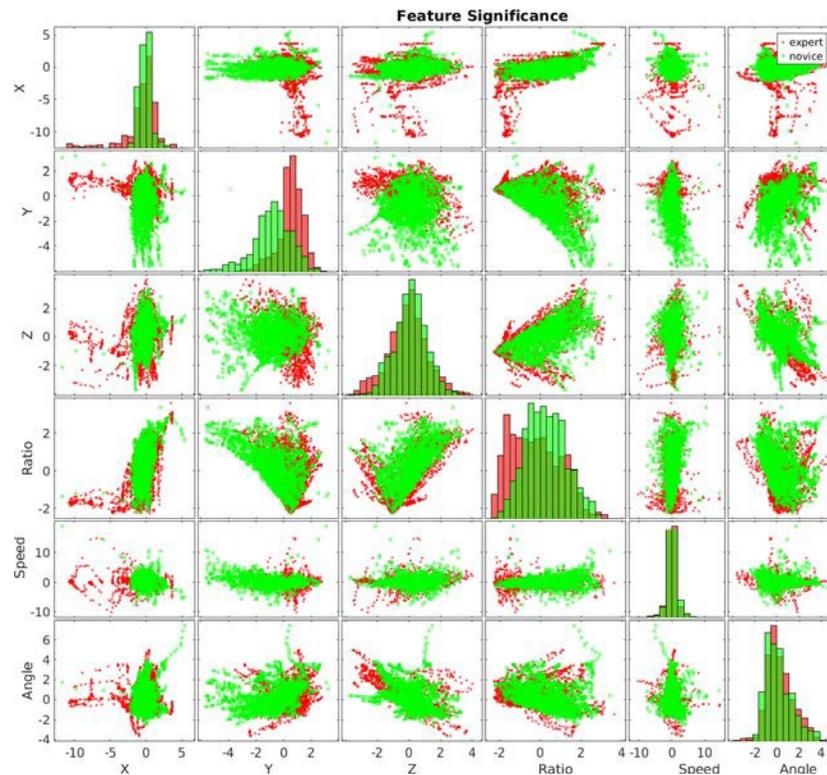

Fig. 6: Experts(in green) and novices (in red) show significant differences in hand movements.